\documentclass[twoside,11pt]{article}

\usepackage{automl2020}

\usepackage{booktabs}
\usepackage{subcaption}
\usepackage{bm}
\usepackage{color}
\usepackage{listings}       

\usepackage{graphicx}
\usepackage{amsmath}

\usepackage[noend]{algpseudocode}
\usepackage{algorithm}
\usepackage{tabularx}

\usepackage{xcolor}
\usepackage{placeins}
\usepackage{adjustbox}

\newcommand{\multiline}[1]{%
  \begin{tabularx}{\dimexpr\linewidth}[t]{@{}X@{}}
    #1
  \end{tabularx}
}

\jmlrheading{A. Devos and M. Grossglauser}

\ShortHeadings{Regression Networks for Meta-Learning Few-Shot Classification}{Devos and Grossglauser}
\firstpageno{1}

\begin{document}

\title{Regression Networks for Meta-Learning Few-Shot Classification}

\author{\name Arnout Devos \email arnout.devos@epfl.ch
        \vspace{-.2cm}
       \AND
       \name Matthias Grossglauser \email matthias.grossglauser@epfl.ch
       \\
       \addr \'Ecole Polytechnique F\'ed\'erale de Lausanne (EPFL), Switzerland
       }

\maketitle

\begin{abstract}
We propose \textit{regression networks} for the problem of few-shot classification, where a classifier must generalize to new classes not seen in the training set, given only a small number of examples of each class. In high dimensional embedding spaces the direction of data generally contains richer information than magnitude. Next to this, state-of-the-art few-shot metric methods that compare distances with aggregated class representations, have shown superior performance. Combining these two insights, we propose to meta-learn classification of embedded points by regressing the closest approximation in every class subspace while using the regression error as a distance metric.
Similarly to recent approaches for few-shot learning, regression networks reflect a simple inductive bias that is beneficial in this limited-data regime and they achieve excellent results, especially when more aggregate class representations can be formed with multiple shots. The code for this project is publicly available at \href{https://github.com/ArnoutDevos/RegressionNet}{\texttt{https://github.com/ArnoutDevos/RegressionNet}}
\end{abstract}

\section{Introduction}
The ability to adapt quickly to new situations is a cornerstone of human intelligence. Artificial learning methods have been shown to be very effective for specific tasks, often surpassing human performance \citep{alphago,dermato}. However, by relying on standard training paradigms for supervised learning or reinforcement learning, these artificial methods still require much training data and training time to adapt to a new task.

An area of machine learning that learns and adapts from a small amount of data is called \textit{few-shot learning} \citep{fei2006one}. A \textit{shot} corresponds to a single example, e.g., an image and its label. In few-shot learning the learning scope is expanded from the classic setting of a single task with many shots to a variety of tasks with a few shots each. Several machine learning approaches have been used for this, including metric-learning \citep{vinyals2016matching,snell,sung2018relation}, meta-learning \citep{finn,ravi}, and generative models \citep{fei2006one,lake2015human}.

\begin{figure}
    \centering
    \includegraphics[width=.8\textwidth]{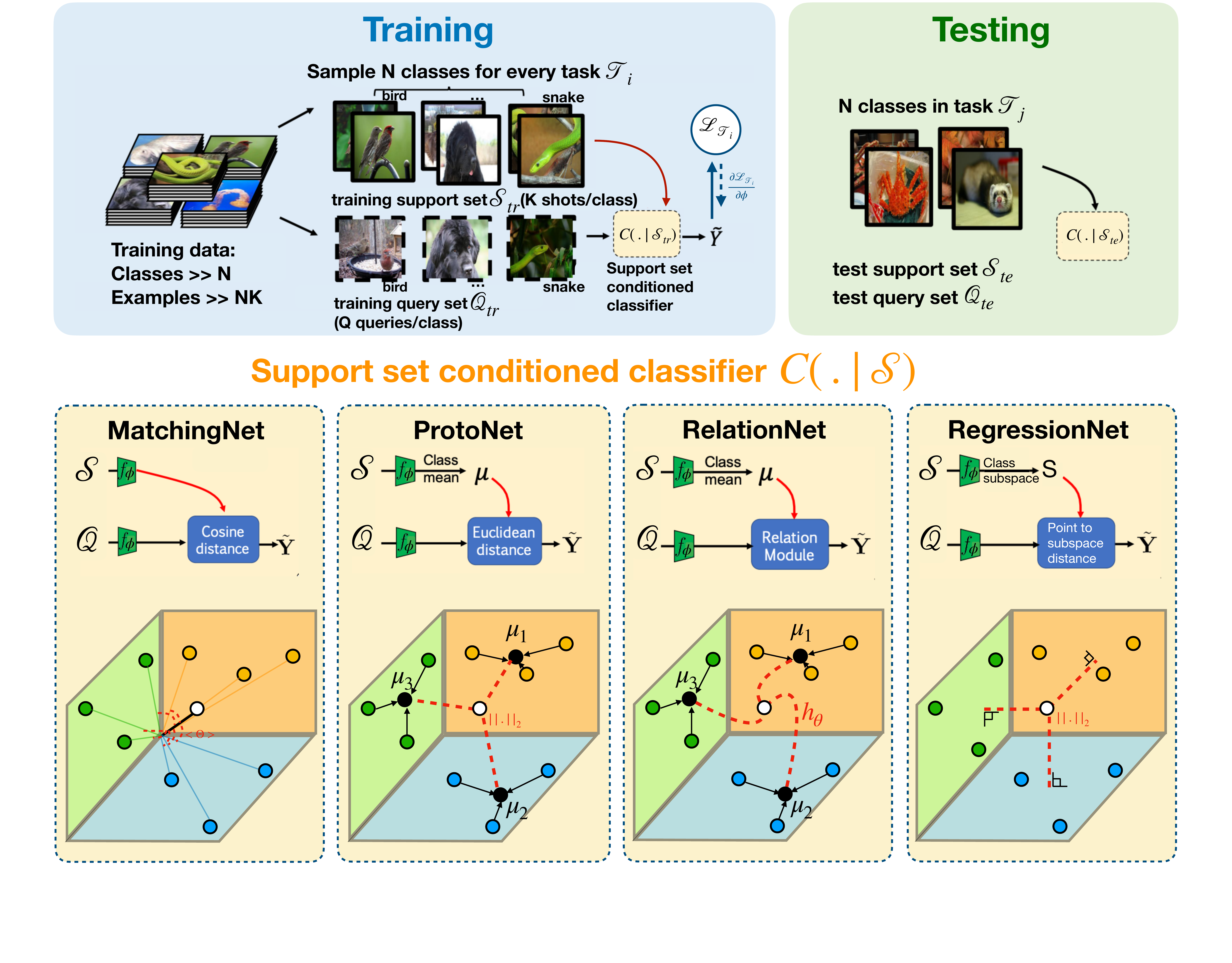}
    \vspace{-1cm}
    \caption{Few-shot learning process (top) and metric-learning based methods (bottom), with $\mathcal{S}$ the support set (colored circles), $\mathcal{Q}$ the query set (white circle), $\widetilde{Y}$ the output distribution over classes for the query points, $f_\phi$ a neural embedding function, and $h_\theta$ a neural network based distance function. Figure inspired by \citet{chen2019closer}.\vspace{-.0cm}}
    \label{fig:overview}
\end{figure}
\citet{chen2019closer} show that a metric-learning based method called \textit{prototypical networks} \citep{snell}, although simple in nature, achieves competitive performance with state-of-the-art meta-learning approaches such as MAML \citep{finn} and other metric-learning methods. Metric-learning methods approach the few-shot classification problem by "\textit{learning to compare}". To learn high capacity nonlinear comparison models, most modern few-shot metric-learning methods use a neural embedding space to measure distance.

In image classification with neural embeddings, the embedding dimensions are usually high: e.g., 1600 for a \textit{Conv-4} backbone (used later). In high dimensional image vector spaces "direction" of data generally contains richer information than "magnitude" \citep{zhe2019directional}. MatchingNet \citep{vinyals2016matching} leverages purely directional information whereas ProtoNet \citep{snell} and RelationNet \citep{sung2018relation} mostly improve on this by comparing with aggregated class representations.

We propose \textit{regression networks} which combine the good properties of directional information in high-dimensional vector spaces with rich aggregate class information. The proposed method is based on the idea that there exists an embedding in which every point from the same class can be approximated by a linear combination of other points in that same class. In order to do this, we learn a nonlinear mapping of the input space to an embedding space by using a neural network and regress the best embedded approximation, for each example. Classification of an embedded query point is then performed by simply finding the nearest class subspace by comparing regression errors.

Subspaces have been used to model images for decades in computer vision and machine learning \citep{fitzgibbon2003joint,naseem2010linear}. For example, the linear regression classification (LRC) method \citep{naseem2010linear} relies on the fact that the set of all reflectance functions produced by Lambertian objects, which parts of natural images are composed of, lie near a low-dimensional vector subspace \citep{basri2003lambertian}. More recently, \citet{simon2019projective} have explored few-shot learning with affine subspaces. Unlike our approach with vector subspaces, affine subspaces cannot be constructed with 1-shot learning, a key few-shot learning problem.

Figure \ref{fig:overview} shows an overview of the few-shot learning process and a comparison of the proposed method with comparable state-of-the-art metric-learning based approaches.

\section{Regression Networks}
\subsection{Notation}
\newcommand{\norm}[1]{\left\lVert#1\right\rVert}
We formulate the $N$-way $K$-shot classification problem in an episodic way. Every episode has a small support set of $N$ classes with $K$ labeled examples $\mathcal{S} = \{(\mathbf{x}_{11}, y_{11}),\ldots,(\mathbf{x}_{NK}, y_{NK})\}$, and a query set of $Q$ different examples $\mathcal{Q} = \{(\mathbf{x}_{1(K+1)}, y_{1(K+1)}),\ldots,(\mathbf{x}_{N(K+Q)}, y_{N(K+Q)})\}$. Note that the query set contains labels only during training, and the goal is to predict the labels of the query set during testing. In $\mathcal{S}$ and $\mathcal{Q}$ each $\mathbf{x}_{ij} \in \mathbb{R}^D$ is the $D$-dimensional feature vector of an example and $y_{ij} \in \{1,\ldots,N\}$ is the corresponding label.
\subsection{Model}
Regression networks perform classification by regressing the best approximation to an embedding point in each aggregated class representation and subsequently using the regression error as a measure of distance. For this we start by constructing a $K$-dimensional embedded subspace $\mathbf{S}_n$ of each class $n$, given $K$ shots per class, through an embedding function $f_{\phi}:\mathbb{R}^D\rightarrow \mathbb{R}^M$ with learnable parameters $\phi$.
With slight abuse of notation, every class is represented by its subspace matrix $\mathbf{S}_n \in \mathbb{R}^{M \times K}$, that contains the $K$ vectors of the embedded class support points:
\begin{align}
    \mathbf{S}_n = \begin{bmatrix}f_{\phi}(\mathbf{x}_{n1})& \ldots &  f_{\phi}(\mathbf{x}_{nK})
    \end{bmatrix}
\end{align}

The \textit{regression-error distance} $\widetilde{d}(\mathbf{e}_i,\mathbf{S}_n)$ of a point $\mathbf{e}_i \in \mathbb{R}^M$ in the embedding space to a class subspace $\mathbf{S}_n$ can be measured by regressing the closest point in the subspace in terms of a certain distance metric. The closest point can be constructed with a linear combination (represented by vector $\mathbf{a} \in \mathbb{R}^{K\times 1}$) of the embedded support examples spanning that space. By using a Euclidean distance metric, this can be formulated as a quadratic optimization problem in $\mathbf{a}$ of the following form:
\begin{equation}
    \widetilde{d}(\mathbf{e}_i,\mathbf{S}_n) = \underset{\mathbf{a}}{\min} \ d(\mathbf{e}_i,\mathbf{S}_n\mathbf{a}) = \underset{\mathbf{a}}{\min} \norm{\mathbf{e}_i - \mathbf{S}_n\mathbf{a}}_2
    \label{eq:lincomb}
\end{equation}
The associated learning problem with this few-shot ($K$) high-dimensional ($M$) overdeterminded system is \textit{least-squares linear regression}. Given that $M \geq K$, this is a well conditioned system, and it admits a \textit{differentiable} closed-form solution \citep{friedman2001elements}:
\begin{align}
    \widetilde{d}(\mathbf{e}_i,\mathbf{S}_n) &= \norm{\mathbf{e}_i - \mathbf{S}_n\left(\mathbf{S}_n^T \mathbf{S}_n\right)^{-1}\mathbf{S}_n^T\mathbf{e}_i}_2
    \label{eq:closed-form}\\
    &= \norm{\mathbf{e}_i - \mathbf{P}_n\mathbf{e}_i}_2
\end{align}
where it is important to note that the matrix to be inverted is of size $K\times K$ and the number of shots $K$ is usually small.
The transformation matrix $\mathbf{P}_n$ projects the point $\mathbf{e}_i$ orthogonally onto the subspace spanned by the columns of $\mathbf{S}_n$ \citep{kocc2014application}. Note that this transformation matrix $\mathbf{P}_n$ has to be computed only for every class, not every example, which speeds up practical computation. Because the embedding function $f_\phi$ can output linearly dependent embeddings for different support examples, a small term $\lambda_1 > 0$ is added to avoid singularity:
\begin{equation}
    \widetilde{d}(\mathbf{e}_i,\mathbf{S}_n) = \norm{\mathbf{e}_i - \mathbf{S}_n\left(\mathbf{S}_n^T \mathbf{S}_n + \lambda_1\mathbf{I}\right)^{-1}\mathbf{S}_n^T\mathbf{e}_i}_2
    \label{eq:conditioning}
\end{equation}

With this point-to-subspace distance function $\widetilde{d} : \mathbb{R}^{M} \times \mathbb{R}^{M\times K} \rightarrow [0,+\infty)$, regression networks give a distribution over classes for a query point $\mathbf{x}$ based on a softmax over distances to each of the class subspaces in the embedding space:
\begin{align}
    p_{\phi}(y=n\mid \mathbf{x})=\frac{\exp\left(-\widetilde{d}(f_{\phi}(\mathbf{x}),\mathbf{S}_n)\right)}{\sum_{n'}\exp\left(-\widetilde{d}(f_{\phi}(\mathbf{x}),\mathbf{S}_{n'})\right)}
\end{align}

Meta-learning continues by minimizing the negative log-probability function $\mathcal{L}_{base}(\phi) = -\log p_{\phi}(y=n\mid \mathbf{x})$ of the true class $n$ via stochastic gradient descent (SGD). Training episodes are formed by randomly sampling a subset of $N$ classes from the training set. Then, a subset of $K$ examples within each class is chosen as the support set $\mathcal{S}$, and a subset of $Q$ examples within each class is chosen as the query set $\mathcal{Q}$.

\subsection{Subspace Orthogonalization}\label{sec:orthogonalization}
The $K$-dimensional class vector subspaces live in a much larger $M$-dimensional space. Therefore, we can exploit this freedom during training by making subspaces as directionally different as possible. Concretely, we propose to add a pairwise subspace orthogonolization term to the loss function:
\begin{equation}
    \mathcal{L}_T = \mathcal{L}_{base} + \lambda_2\sum^N_{i\neq j}\frac{\norm{S_i^T S_j}^2_F}{\norm{S_i}^2_F\norm{S_j}^2_F}
    \label{eq:orthogonalization}
\end{equation}
where $\norm{\cdot}_F$ is the Frobenius norm and $\lambda_2$ is a scaling hyperparameter. Section \ref{sec:ablation} studies the effect of this term. We have also experimented with principal angles between vector subspaces \citep{bjorck1973numerical}, because they only depend on the subspaces, not on the (non-unique) set of points that define the subspaces as in Equation \eqref{eq:orthogonalization}. Their results were comparable with our current approach, but come at a higher computational cost with a singular value decomposition.

Algorithm \ref{alg:regressionnet} in Appendix \ref{app:algorithm} details the complete regression networks training procedure.

\section{Experiments}
In terms of few-shot learning evaluation, we focus on the natural image-based mini-ImageNet \citep{vinyals2016matching} dataset. To ensure a fair comparison with other methods, we perform experiments under the same conditions using the verified re-implementation \citep{chen2019closer} of MatchingNet, ProtoNet, RelationNet, MAML and extend it with R2D2 \citep{bertinetto2018metalearning}. Compared to our direct approach, R2D2 is a meta-learning technique which leverages the closed-form solution of multinomial regression indirectly for classification (See Section \ref{sec:relatedwork}). We decide to compare with these methods in particular, because they serve as the basis of many state-of-the-art few-shot classification algorithms \citep{oreshkin2018tadam,Xing2019adaptive}, and our method is easily interchanged with them. Experimental details can be found in Appendix \ref{app:implementation}.

In this section, next to performance evaluation, we address the following research questions: (i) Can regression networks benefit from richer class representations and higher dimensions? (Section \ref{sec:classification}). (ii) How much effect does subspace orthogonalization have? (Section \ref{sec:ablation}).

\subsection{Few-shot Image Classification: mini-ImageNet}\label{sec:classification}
Table \ref{tab:miniconv4} shows the results for 5-way classification for mini-ImageNet for a different number of shots and backbones.

\begin{table}
  \centering
  \begin{adjustbox}{width=\columnwidth,center}
  \begin{tabular}{l c c c c}
    \toprule
    &\multicolumn{2}{c}{\bf Conv-4} & \multicolumn{2}{c}{\bf ResNet-10}             \\
    \cmidrule(r){2-5}
    {\bf Method }   & {\bf 5-way 1-shot} & {\bf 5-way 5-shot} & {\bf 5-way 1-shot} & {\bf 5-way 5-shot} \\
    \midrule
    MatchingNet      & 48.14$\pm$0.78 & 63.28$\pm$0.68 & 54.49$\pm$0.81 & 69.14$\pm$0.69\\
    ProtoNet       & 44.42$\pm$0.84 & 65.15$\pm$0.67 &  51.98$\pm$0.84 & 73.77$\pm$0.64\\
    RelationNet    & 49.31$\pm$0.85 & 65.33$\pm$0.70 &  52.19$\pm$0.83 & 69.97$\pm$0.68\\
    MAML    & 46.47$\pm$0.82 & 62.71$\pm$0.71 &  54.69$\pm$0.89 &  66.62$\pm$0.83\\
    R2D2    & \textbf{50.07}$\pm$0.79 & 65.66$\pm$0.69 & \textbf{55.71}$\pm$0.78 & 71.69$\pm$0.63\\
    RegressionNet (ours)    & 47.02$\pm$0.77 & \textbf{67.09}$\pm$0.69 & 55.44$\pm$0.86 & \textbf{76.29}$\pm$0.59\\
    \bottomrule
    
  \end{tabular}
  \end{adjustbox}
  \caption{Average accuracies (\%) of mini-ImageNet test tasks with 95\% confidence intervals.
  }
  \label{tab:miniconv4}
\end{table}

First, because regression networks are expected to benefit more from better subspace representations when more support examples are available per class, we investigate the effect of the number of shots. As expected, when increasing the number of shots $K$ per class from 1 to 5, the classification accuracies increase for all methods. In the 5-shot case, RegressionNet significantly outperforms all other methods, showing the benefit of using rich class representations.

Secondly, as the backbone gets deeper, regression networks and prototypical networks begin to perform significantly better than matching networks and relation networks with R2D2 following close. Although the performance difference is small for mini-Imagenet, given a relatively deep feature extractor ResNet-10, regression networks outperform all other meta-learning and metric-learning methods when enough shots are available.

\subsection{Ablation study}\label{sec:ablation}
In order to evaluate the effect of adding the subspace orthogonalization stimulating term to the loss function discussed in Section \ref{sec:orthogonalization}, we conduct an experiment without it. The results, comparing a ResNet-10 model trained with subspace orthogonalization and without, are shown in Table \ref{tab:ablation}. All ablation experiments use ResNet-10 as a backbone.

Under all settings considered, subspace orthogonalization gives a classification accuracy improvement (up to 2\%). Note that, even without subspace orthogonalization, our proposed method is still competitive with all other methods in Table \ref{tab:miniconv4}.
\begin{table}
  \centering
  \begin{tabular}{l c c}
    \toprule
    $N$-way $K$-shot & $\lambda_2=0$ & $\lambda_2 > 0$\\
    \midrule
    5-way 1-shot  & 54.83$\pm$0.83   &   55.44$\pm$0.86 \\
    5-way 5-shot  & 74.03$\pm$0.68   &   76.29$\pm$0.59 \\
    \bottomrule
  \end{tabular}
  \caption{Ablation study of effect of subspace orthogonalization stimulation (Equation \eqref{eq:orthogonalization}) using a ResNet-10 backbone on mini-ImageNet. 1-shot: $\lambda_2 = 10^{-3}$, 5-shot: $\lambda_2 = 10^{-2}$ }
  \label{tab:ablation}
\end{table}

\section{Related work}\label{sec:relatedwork}
In addition to the reproduced metric (meta-)learning based few-shot methods \citep{snell,vinyals2016matching,sung2018relation, bertinetto2018metalearning}, there is a large body of work on few-shot learning and metric (meta-)learning. We discuss work that is more closely related to regression networks in particular.

Our approach shows similarities to the linear regression classification (LRC) method \citep{naseem2010linear}, where each class is represented by the vector subspace spanned by its examples. LRC was developed for face recognition, where only a few examples are available, however it relies on a linear embedding. Our approach also uses few examples, but it incorporates neural networks in order to nonlinearly embed examples and we couple this with episodic training to handle the meta-learning few-shot scenario.

\citet{simon2019projective} have explored affine subspace representations for few-shot learning. In contrast to our closed-form linear regression approach, they make use of a truncated singular value decomposition (SVD) of the support example matrix. Affine subspaces cannot be constructed with 1-shot learning, a key few-shot learning problem. In contrast, our closed-form linear regression approach relies on vector subspaces, which can be constructed with 1-shot learning.

\citet{bertinetto2018metalearning} propose to use regularized linear regression as a classifier on top of the embedding function. Doing so, they directly approach a classification problem with a regression method, but they show competitive results. To achieve this, they introduce learnable scalars that scale and shift the regression outputs for them to be used in the cross-entropy loss. Regression networks rely on the same closed-form solver for linear regression to compute the transformation matrices, but are inherently designed for classification problems because of their similarity to the LRC method \citep{naseem2010linear}.

\cite{oreshkin2018tadam} show that, following \cite{hinton2015distilling}, by adding a learnable temperature $\alpha$ in the softmax of metric-based few-shot models (e.g., for regression networks: $p_{\phi,\alpha}(y=n|\mathbf{x})=\text{softmax}(-\alpha d(\mathbf{e}_i,\mathbf{P}_n\mathbf{e}_i))$), the performance difference between matching networks and prototypical networks vanishes. However, the performance of prototypical networks improves only slightly. In addition, they make the embedding function conditionally dependent on the support set with conditional batch normalization. They show that by combining support-set conditioned batch normalization with temperature learning, impressive performance gains can be achieved.

\section{Conclusion}
We have proposed regression networks for meta-learning few-shot classification. The method assumes that for any embedded point we can regress the closest approximation in every class representation and use the error as a distance measure. Classes are represented by their embedded vector subspaces, which are spanned by their examples. The approach produces better results than other state-of-the-art metric-learning based methods, when rich class representations can be formed with multiple shots. Stimulating subspace orthogonality consistently improves performance. A direction for future work is to study the effect of using a low-rank approximation of the class subspace. Overall, the simplicity and effectiveness of regression networks makes it a promising approach for metric-based few-shot classification.

\section*{Acknowledgements}
We acknowledge fruitful discussions with Andreas Loukas, William Trouleau, Gergely Odor, Farnood Salehi, Negar Foroutan and anonymous reviewers. This project is supported by the European Union's Horizon 2020 research and innovation program under the Marie Skłodowska-Curie grant agreement No. 754354.

\vskip 0.2in
\bibliography{automl2020}

\begin{thebibliography}{29}
\providecommand{\natexlab}[1]{#1}
\providecommand{\url}[1]{\texttt{#1}}
\expandafter\ifx\csname urlstyle\endcsname\relax
  \providecommand{\doi}[1]{doi: #1}\else
  \providecommand{\doi}{doi: \begingroup \urlstyle{rm}\Url}\fi

\bibitem[Basri and Jacobs(2003)]{basri2003lambertian}
Ronen Basri and David~W Jacobs.
\newblock Lambertian reflectance and linear subspaces.
\newblock \emph{IEEE Transactions on Pattern Analysis \& Machine Intelligence},
  \penalty0 (2):\penalty0 218--233, 2003.

\bibitem[{Bertinetto} et~al.(2019){Bertinetto}, {Henriques}, {Torr}, and
  {Vedaldi}]{bertinetto2018metalearning}
Luca {Bertinetto}, João~F. {Henriques}, Philip H.~S. {Torr}, and Andrea
  {Vedaldi}.
\newblock Meta-learning with differentiable closed-form solvers.
\newblock In \emph{ICLR 2019 : 7th International Conference on Learning
  Representations}, 2019.
\newblock URL \url{https://academic.microsoft.com/paper/2964206659}.

\bibitem[Bjorck and Golub(1973)]{bjorck1973numerical}
Ake Bjorck and Gene~H Golub.
\newblock Numerical methods for computing angles between linear subspaces.
\newblock \emph{Mathematics of computation}, 27\penalty0 (123):\penalty0
  579--594, 1973.

\bibitem[Chen et~al.(2019)Chen, Liu, Kira, Wang, and Huang]{chen2019closer}
Wei-Yu Chen, Yen-Cheng Liu, Zsolt Kira, Yu-Chiang~Frank Wang, and Jia-Bin
  Huang.
\newblock A closer look at few-shot classification.
\newblock In \emph{International Conference on Learning Representations}, 2019.

\bibitem[Deng et~al.(2009)Deng, Dong, Socher, Li, Li, and
  Fei-Fei]{deng2009imagenet}
Jia Deng, Wei Dong, Richard Socher, Li-Jia Li, Kai Li, and Li~Fei-Fei.
\newblock Imagenet: A large-scale hierarchical image database.
\newblock In \emph{2009 IEEE conference on computer vision and pattern
  recognition}, pages 248--255. Ieee, 2009.

\bibitem[Esteva et~al.(2017)Esteva, Kuprel, Novoa, Ko, Swetter, Blau, and
  Thrun]{dermato}
Andre Esteva, Brett Kuprel, Roberto~A Novoa, Justin Ko, Susan~M Swetter,
  Helen~M Blau, and Sebastian Thrun.
\newblock Dermatologist-level classification of skin cancer with deep neural
  networks.
\newblock \emph{Nature}, 542\penalty0 (7639):\penalty0 115, 2017.

\bibitem[Fei-Fei et~al.(2006)Fei-Fei, Fergus, and Perona]{fei2006one}
Li~Fei-Fei, Rob Fergus, and Pietro Perona.
\newblock One-shot learning of object categories.
\newblock \emph{IEEE transactions on pattern analysis and machine
  intelligence}, 28\penalty0 (4):\penalty0 594--611, 2006.

\bibitem[{Finn} et~al.(2017){Finn}, {Abbeel}, and {Levine}]{finn}
Chelsea {Finn}, Pieter {Abbeel}, and Sergey {Levine}.
\newblock Model-agnostic meta-learning for fast adaptation of deep networks.
\newblock In \emph{ICML'17 Proceedings of the 34th International Conference on
  Machine Learning - Volume 70}, pages 1126--1135, 2017.
\newblock URL \url{https://academic.microsoft.com/paper/2604763608}.

\bibitem[Fitzgibbon and Zisserman(2003)]{fitzgibbon2003joint}
Andrew~W Fitzgibbon and Andrew Zisserman.
\newblock Joint manifold distance: a new approach to appearance based
  clustering.
\newblock In \emph{2003 IEEE Computer Society Conference on Computer Vision and
  Pattern Recognition, 2003. Proceedings.}, volume~1, pages I--I. IEEE, 2003.

\bibitem[Friedman et~al.(2001)Friedman, Hastie, and
  Tibshirani]{friedman2001elements}
Jerome Friedman, Trevor Hastie, and Robert Tibshirani.
\newblock \emph{The elements of statistical learning}, volume~1.
\newblock Springer series in statistics New York, 2001.

\bibitem[He et~al.(2016)He, Zhang, Ren, and Sun]{he2016deep}
Kaiming He, Xiangyu Zhang, Shaoqing Ren, and Jian Sun.
\newblock Deep residual learning for image recognition.
\newblock In \emph{Proceedings of the IEEE conference on computer vision and
  pattern recognition}, pages 770--778, 2016.

\bibitem[Hilliard et~al.(2018)Hilliard, Phillips, Howland, Yankov, Corley, and
  Hodas]{hilliard2018few}
Nathan Hilliard, Lawrence Phillips, Scott Howland, Art{\"e}m Yankov, Courtney~D
  Corley, and Nathan~O Hodas.
\newblock Few-shot learning with metric-agnostic conditional embeddings.
\newblock \emph{arXiv preprint arXiv:1802.04376}, 2018.

\bibitem[Hinton et~al.(2015)Hinton, Vinyals, and Dean]{hinton2015distilling}
Geoffrey Hinton, Oriol Vinyals, and Jeff Dean.
\newblock Distilling the knowledge in a neural network.
\newblock \emph{arXiv preprint arXiv:1503.02531}, 2015.

\bibitem[Kingma and Ba(2014)]{kingma2014adam}
Diederik~P Kingma and Jimmy Ba.
\newblock Adam: A method for stochastic optimization.
\newblock \emph{arXiv preprint arXiv:1412.6980}, 2014.

\bibitem[Ko{\c{c}} and Barkana(2014)]{kocc2014application}
Mehmet Ko{\c{c}} and Atalay Barkana.
\newblock Application of linear regression classification to low-dimensional
  datasets.
\newblock \emph{Neurocomputing}, 131:\penalty0 331--335, 2014.

\bibitem[Lake et~al.(2015)Lake, Salakhutdinov, and Tenenbaum]{lake2015human}
Brenden~M Lake, Ruslan Salakhutdinov, and Joshua~B Tenenbaum.
\newblock Human-level concept learning through probabilistic program induction.
\newblock \emph{Science}, 350\penalty0 (6266):\penalty0 1332--1338, 2015.

\bibitem[Naseem et~al.(2010)Naseem, Togneri, and Bennamoun]{naseem2010linear}
Imran Naseem, Roberto Togneri, and Mohammed Bennamoun.
\newblock Linear regression for face recognition.
\newblock \emph{IEEE transactions on pattern analysis and machine
  intelligence}, 32\penalty0 (11):\penalty0 2106--2112, 2010.

\bibitem[Oreshkin et~al.(2018)Oreshkin, L{\'o}pez, and
  Lacoste]{oreshkin2018tadam}
Boris Oreshkin, Pau~Rodr{\'\i}guez L{\'o}pez, and Alexandre Lacoste.
\newblock Tadam: Task dependent adaptive metric for improved few-shot learning.
\newblock In \emph{Advances in Neural Information Processing Systems}, pages
  721--731, 2018.

\bibitem[Paszke et~al.(2017)Paszke, Gross, Chintala, Chanan, Yang, DeVito, Lin,
  Desmaison, Antiga, and Lerer]{paszke2017automatic}
Adam Paszke, Sam Gross, Soumith Chintala, Gregory Chanan, Edward Yang, Zachary
  DeVito, Zeming Lin, Alban Desmaison, Luca Antiga, and Adam Lerer.
\newblock Automatic differentiation in pytorch.
\newblock 2017.

\bibitem[Qiao et~al.(2018)Qiao, Liu, Shen, and Yuille]{qiao2018few}
Siyuan Qiao, Chenxi Liu, Wei Shen, and Alan~L Yuille.
\newblock Few-shot image recognition by predicting parameters from activations.
\newblock In \emph{Proceedings of the IEEE Conference on Computer Vision and
  Pattern Recognition}, pages 7229--7238, 2018.

\bibitem[{Ravi} and {Larochelle}(2017)]{ravi}
Sachin {Ravi} and Hugo {Larochelle}.
\newblock Optimization as a model for few-shot learning.
\newblock In \emph{ICLR 2017 : International Conference on Learning
  Representations 2017}, 2017.
\newblock URL \url{https://academic.microsoft.com/paper/2753160622}.

\bibitem[Silver et~al.(2016)Silver, Huang, Maddison, Guez, Sifre, Van
  Den~Driessche, Schrittwieser, Antonoglou, Panneershelvam, Lanctot,
  et~al.]{alphago}
David Silver, Aja Huang, Chris~J Maddison, Arthur Guez, Laurent Sifre, George
  Van Den~Driessche, Julian Schrittwieser, Ioannis Antonoglou, Veda
  Panneershelvam, Marc Lanctot, et~al.
\newblock Mastering the game of go with deep neural networks and tree search.
\newblock \emph{nature}, 529\penalty0 (7587):\penalty0 484, 2016.

\bibitem[Simon et~al.(2018)Simon, Koniusz, and Harandi]{simon2019projective}
Christian Simon, Piotr Koniusz, and Mehrtash Harandi.
\newblock Projective subspace networks for few-shot learning, 2018.
\newblock URL \url{https://openreview.net/forum?id=rkzfuiA9F7}.

\bibitem[{Snell} et~al.(2017){Snell}, {Swersky}, and {Zemel}]{snell}
Jake {Snell}, Kevin {Swersky}, and Richard~S. {Zemel}.
\newblock Prototypical networks for few-shot learning.
\newblock In \emph{Advances in Neural Information Processing Systems}, pages
  4077--4087, 2017.
\newblock URL \url{https://academic.microsoft.com/paper/2601450892}.

\bibitem[Sung et~al.(2018)Sung, Yang, Zhang, Xiang, Torr, and
  Hospedales]{sung2018relation}
Flood Sung, Yongxin Yang, Li~Zhang, Tao Xiang, Philip~HS Torr, and Timothy~M
  Hospedales.
\newblock Learning to compare: Relation network for few-shot learning.
\newblock In \emph{Proceedings of the IEEE Conference on Computer Vision and
  Pattern Recognition}, pages 1199--1208, 2018.

\bibitem[Vinyals et~al.(2016)Vinyals, Blundell, Lillicrap, Wierstra,
  et~al.]{vinyals2016matching}
Oriol Vinyals, Charles Blundell, Timothy Lillicrap, Daan Wierstra, et~al.
\newblock Matching networks for one shot learning.
\newblock In \emph{Advances in neural information processing systems}, pages
  3630--3638, 2016.

\bibitem[Wah et~al.(2011)Wah, Branson, Welinder, Perona, and
  Belongie]{wah2011caltech}
Catherine Wah, Steve Branson, Peter Welinder, Pietro Perona, and Serge
  Belongie.
\newblock The caltech-ucsd birds-200-2011 dataset.
\newblock 2011.

\bibitem[Xing et~al.(2019)Xing, Rostamzadeh, Oreshkin, and
  Pinheiro]{Xing2019adaptive}
Chen Xing, Negar Rostamzadeh, Boris~N. Oreshkin, and Pedro~O. Pinheiro.
\newblock Adaptive cross-modal few-shot learning.
\newblock \emph{CoRR}, abs/1902.07104, 2019.
\newblock URL \url{http://arxiv.org/abs/1902.07104}.

\bibitem[Zhe et~al.(2019)Zhe, Chen, and Yan]{zhe2019directional}
Xuefei Zhe, Shifeng Chen, and Hong Yan.
\newblock Directional statistics-based deep metric learning for image
  classification and retrieval.
\newblock \emph{Pattern Recognition}, 93:\penalty0 113--123, 2019.

\end{thebibliography}


\newpage

\appendix
\section{Algorithm}\label{app:algorithm}
\begin{algorithm*}
 \caption{Regression Networks}
 \begin{algorithmic}[1]
 \Require Training set with task batches of $N_T$ $N$-way $K$-shot episodes/tasks $\{{\mathcal{T}_1,\ldots,\mathcal{T}_{N_T}}\}$, with every $\mathcal{T}_i$ containing sets $\mathcal{S}_{\mathcal{T}_i} = \{(\mathbf{x}_{i11}, y_{i11}),\ldots,(\mathbf{x}_{iNK}, y_{iNK})\}$ and $\mathcal{Q}_{\mathcal{T}_i} = \{(\mathbf{x}_{i1(K+1)}, y_{i1(K+1)}),\ldots,(\mathbf{x}_{iN(K+Q)}, y_{iN(K+Q)})\}$. $\mathcal{S}_{\mathcal{T}_i n}$ and $\mathcal{Q}_{\mathcal{T}_i n}$ denote the class $n$ subsets of support and query sets, respectively, of episode $\mathcal{T}_i$.
 \Require $\alpha, \lambda_1, \lambda_2$: step size, conditioning and orthogonalization parameters
 \State Randomly initialize $\phi$
 \While {not done}
 \For {each batch} \Comment{Sample batch of tasks}
  \For {$i$ \textbf{in} $\{1,\ldots,N_T\}$}\Comment{Select task}
   
   \For {n \textbf{in} $\{1, \ldots, N\}$} \Comment{Select class}

    \State $\mathbf{S}_n \leftarrow f_{\phi}(\mathcal{S}_{\mathcal{T}_i n})$ \Comment{Embed class support set subspace $\mathbf{S}_n \in \mathbb{R}^{M\times K}$}
    \State $\mathbf{P}_n \leftarrow \mathbf{S}_n\left(\mathbf{S}_n^T \mathbf{S}_n + \lambda_1\mathbf{I}\right)^{-1}\mathbf{S}_n^T$ \Comment{Compute transformation matrix}
   \EndFor
   \State $\mathcal{L}_{\mathcal{T}_i} \leftarrow 0$ \Comment{Initialize episode loss}
   \For {$n$ \textbf{in} $\{1, \ldots, N\}$}

    \For {$(\mathbf{x},y=n)$ \textbf{in} $\mathcal{Q}_{\mathcal{T}_i n}$}

     \State \multiline{%
     $\mathcal{L}_{\mathcal{T}_i} \hspace*{-1mm} \leftarrow \hspace*{-1mm} \mathcal{L}_{\mathcal{T}_i} +  \frac{1}{NQ}\bigg[\hspace*{-1mm}\norm{f_{\phi}(\mathbf{x}) - \mathbf{P}_{n} f_{\phi}(\mathbf{x})}_2 + \log{\sum\limits_{n'} \exp(-\norm{f_{\phi}(\mathbf{x}) - \mathbf{P}_{n'} f_{\phi}(\mathbf{x})}_2})\bigg] \newline + \lambda_2\sum^N_{i\neq j}\frac{\norm{\mathbf{S}_i^T \mathbf{S}_j}^2_F}{\norm{\mathbf{S}_i}^2_F\norm{\mathbf{S}_j}^2_F}$}
    \EndFor
   \EndFor
  \EndFor
  \State $\phi \leftarrow \phi - \alpha\nabla_\phi \sum\limits_i\mathcal{L}_{\mathcal{T}_i}$ \Comment{Update embedding parameters $\phi$ with gradient descent}
  \EndFor
 \EndWhile
 \end{algorithmic}
 \label{alg:regressionnet}
\end{algorithm*}

\section{Experimental details}\label{app:implementation}

\subsection{Experimental Setup and Datasets}

The mini-Imagenet dataset proposed by \citep{vinyals2016matching} contains 100 classes, with 600 84 $\times$ 84 images per class sampled from the larger ImageNet dataset \citep{deng2009imagenet}. Following \citet{ravi}, 64 classes are isolated for the training set and, from the remaining classes, the validation and test sets of 16 and 20 classes, respectively, are constructed. We use exactly the same train/validation/test split of classes as the one suggested by \citet{ravi}.

For the additional experiments in Appendix \ref{app:experiments} and cross-domain experiments in Appendix \ref{app:cross} we additionaly make use of the CUB-200-2011 dataset \citep{wah2011caltech}, hereafter referred to as \textit{CUB}. The CUB dataset contains 200 classes of birds and 11,788 images in total. Following \citet{hilliard2018few}, we randomly split the CUB dataset into 100 train, 50 validation, and 50 test classes.

We implement regression networks using the automatic-differentation framework PyTorch \citep{paszke2017automatic}.

Many training optimizations exist, including using more classes in the training episodes than in the testing episodes \citep{snell} or pre-training the feature extractor with all training classes and a linear output layer \citep{qiao2018few}. However, we train all methods from scratch and construct our training and testing episodes to have the same number of classes $N$ and shots $K$ because we are interested in the relative performance of the methods. For RegressionNet, the conditioning parameter used to ensure a fully invertible matrix in Equation \eqref{eq:conditioning} is set to $\lambda_1=10^{-3}$. For 1-shot and 5-shot learning, $\lambda_2=10^{-3}$ and $\lambda_2=10^{-2}$ are used, respectively.

\subsection{Architectures and training}
Despite the modification of some implementation details of the methods with respect to the original papers, these settings ensure a fair comparison and \citet{chen2019closer} report a maximal drop in classification performance of 2\% with respect to the original reported performance of each method. For MAML, we reuse the results from \citet{chen2019closer}.

The \textit{Conv-4} backbone is composed of four convolutional blocks with an input size of 84$\times$84 as in \citet{snell}. Each block comprises a 64-filter 3$\times$3 convolution with a padding of 1 and a stride of 1, a batch normalization layer, a ReLU nonlinearity and a 2$\times$2 max-pooling layer.

The \textit{ResNet-10} backbone has an input size of 224$\times$224 and is a simplified version of ResNet-18 in \citet{he2016deep}, by using only one residual building block in each layer.

\textit{ResNet-34} is the same as described in \citet{he2016deep}.

All methods are trained with a random parameter initialization and use the Adam optimizer \citep{kingma2014adam} with an initial learning rate of $10^{-3}$. During the training stage, data augmentation is done in the form of: random crop, color jitter, and left-right flip. For MatchingNet, a rather sophisticated long-short term memory (LSTM)-based full context embedding (FCE) classification layer without fine-tuning is used over the support set, and the cosine similarity metric is multiplied by a constant factor of 100. In RelationNet, the L2 norm is replaced with a softmax layer to help the training procedure \citep{chen2019closer}. The relation module is composed of 2 convolutional blocks (same as in Conv-4), followed by two fully connected layers of size 8 and 1.

In all settings, we train for 60,000 episodes and use the held-out validation set to select the best model during training. To construct an $N$-way episode, we sample $N=5$ classes from the training set of classes. From each sampled class, we sample $K$ examples to construct the support set of an episode and $Q=16$ examples for the query set. During testing, we average the results over 600 episodes of exactly the same form, but this time they are sampled from the test set classes.

\section{Extra experiments: mini-ImageNet and CUB}\label{app:experiments}
We provide additional results for some considered methods on the effect of increasing the feature embedding backbone depth even further and increasing the number of shots to 10. Also, next to the already considered mini-ImageNet few-shot dataset, we also evaluate performance on CUB. Note that for RegressionNet we do not use the subspace orthogonalization regularizer from Section \ref{sec:orthogonalization}. Tables \ref{tab:5way1shotCUBdepth} to \ref{tab:5way10shotminidepth} show the results on this.

As expected, performance of the few-shot classification methods generally improves when a deeper backbone is used. As can be seen, the relative rank in terms of classification accuracy of the distance-metric based methods shifts significantly when going from the Conv-4 embedding architecture to a deeper ResNet-10 or ResNet-34. For example in Table \ref{tab:5way5shotCUBdepth}, RelationNet performs best in the Conv-4 setting, but worst when using ResNet-34 and RegressionNet takes first place. Generally, across datasets and number of shots, regression networks outperform the other considered methods when the backbone is relatively deep (ResNet-10 and deeper).

Since regression networks are expected to build better subspace representations when more support examples are available per class, we investigate the effect of the number of shots. As expected, when increasing the number of shots $K$ per class, the classification accuracies increase for almost all methods. Even for the smaller Conv-4 backbone, regression networks outperform all other methods on both datasets when 10 shots are used.

Comparing the classification results on the CUB and mini-ImageNet datasets, the accuracies are consistently higher for CUB under the same setting (backbone, shots, method). This can be explained by the difference in divergence of classes; this difference is higher in mini-ImageNet than in CUB \citep{chen2019closer}. Appendix \ref{app:cross} discusses extra experiments on the transferrability of different methods under domain-shift and further reports on the difference in divergence between classes in CUB and mini-ImageNet.
\begin{table}
  \centering
  \begin{tabular}{llll}
    \toprule
         & Conv-4     & ResNet-10 & ResNet-34 \\
    \midrule
    MatchingNet      & 61.16 $\pm$ 0.89 & 71.29 $\pm$ 0.90 & 71.44 $\pm$ 0.96 \\
    ProtoNet       & 51.31 $\pm$ 0.91 & 70.13 $\pm$ 0.94  & 72.03 $\pm$ 0.91  \\
    
    RelationNet   & \textbf{62.45} $\pm$ 0.98 & 68.65 $\pm$ 0.91 & 66.20 $\pm$ 0.99 \\
    RegressionNet (ours)   & 59.05 $\pm$ 0.90 & \textbf{72.92} $\pm$ 0.90 &
\textbf{73.74}$\pm$ 0.88 \\ 
    \bottomrule
    
  \end{tabular}
  \caption{CUB 5-way 1-shot classification accuracies with 95\% confidence intervals for different methods and backbones. For each backbone,
the best method is highlighted.}
\label{tab:5way1shotCUBdepth}
\end{table}

\begin{table}
  
  \centering
  \begin{tabular}{llll}
    \toprule
         & Conv-4     & ResNet-10 & ResNet-34 \\
    \midrule
    MatchingNet      & 76.74 $\pm$ 0.67& 83.75  $\pm$ 0.60& 85.96 $\pm$  0.52\\
    ProtoNet       & 76.14 $\pm$ 0.68&  85.70 $\pm$ 0.52& 88.49 $\pm$ 0.46 \\

    RelationNet   & \textbf{78.98} $\pm$ 0.63 & 82.67 $\pm$ 0.61& 84.13 $\pm$  0.57\\
    RegressionNet (ours)   & 78.48 $\pm$ 0.65 & \textbf{87.45} $\pm$ 0.48 &
\textbf{89.30} $\pm$ 0.45   \\ 
    \bottomrule
    
  \end{tabular}
  
  \caption{CUB 5-way 5-shot classification accuracies with 95\% confidence intervals for different methods and backbones. For each backbone,
the best method is highlighted.}
\label{tab:5way5shotCUBdepth}
\end{table}

\begin{table}
  
  \centering
  \begin{tabular}{llll}
    \toprule
         & Conv-4     & ResNet-10 & ResNet-34 \\
    \midrule
    MatchingNet      & 79.27 $\pm$ 0.61 & 85.38 $\pm$ 0.54 & 89.03 $\pm$ 0.46 \\
    ProtoNet       & 81.77 $\pm$ 0.57 & 87.61 $\pm$ 0.44 & 90.35 $\pm$ 0.40\\
    RelationNet    & 82.66 $\pm$ 0.56 & 84.96 $\pm$ 0.53 & 87.29 $\pm$ 0.46 \\
    RegressionNet (ours)    & \textbf{83.02} $\pm$ 0.53 & \textbf{89.28} $\pm$ 0.41 & \textbf{91.46} $\pm$ 0.36\\ 
    \bottomrule
    
  \end{tabular}
  
  \caption{CUB 5-way 10-shot classification accuracies with 95\% confidence intervals for different methods and backbones. For each backbone, the best method is highlighted.}
  \label{tab:5way10shotCUBdepth}
\end{table}

\begin{table}
  
  \centering
  \begin{tabular}{llll}
    \toprule
         & Conv-4     & ResNet-10 & ResNet-34 \\
    \midrule
    MatchingNet      & 48.14 $\pm$ 0.78 & 54.49 $\pm$ 0.81 & 53.20 $\pm$ 0.78 \\
    ProtoNet       & 44.42 $\pm$ 0.84 & 51.98 $\pm$ 0.84 & 53.90 $\pm$ 0.83 \\
    RelationNet    & \textbf{49.31} $\pm$ 0.85 & 52.19 $\pm$ 0.83 & 52.74 $\pm$ 0.83 \\
    RegressionNet (ours)    & 47.99 $\pm$ 0.80 & \textbf{54.83} $\pm$ 0.83 & \textbf{53.93} $\pm$ 0.81\\ 
    \bottomrule
    
  \end{tabular}
  
  \caption{miniImageNet 5-way 1-shot classification accuracies with 95\% confidence intervals for different methods and backbones. For each backbone,
the best method is highlighted.}
\label{tab:5way1shotminidepth}
\end{table}

\begin{table}
  
  \centering
  \begin{tabular}{llll}
    \toprule
         & Conv-4     & ResNet-10 & ResNet-34 \\
    \midrule
    MatchingNet      & 63.28 $\pm$ 0.68 & 69.14 $\pm$ 0.69 & 70.36 $\pm$ 0.70 \\
    ProtoNet       & 65.15 $\pm$ 0.67 & 73.77 $\pm$ 0.64 & 75.14 $\pm$ 0.65 \\
    RelationNet    & 65.33 $\pm$ 0.70 & 69.97 $\pm$ 0.68 & 71.38 $\pm$ 0.68 \\
    RegressionNet (ours)    & \textbf{66.41} $\pm$ 0.66 & \textbf{74.03} $\pm$ 0.68 & \textbf{75.62} $\pm$ 0.61\\ 
    \bottomrule
    
  \end{tabular}
  
  \caption{miniImageNet 5-way 5-shot classification accuracies with 95\% confidence intervals for different methods and backbones. For each backbone,
the best method is highlighted.}
\label{tab:5way5shotminidepth}
\end{table}

\begin{table}
  
  \centering
  \begin{tabular}{llll}
    \toprule
         & Conv-4     & ResNet-10 & ResNet-34 \\
    \midrule
    MatchingNet      & 67.47 $\pm$ 0.64 & 74.63 $\pm$ 0.62 & 73.88  $\pm$ 0.65\\
    ProtoNet       & 72.01 $\pm$ 0.67 & 78.92 $\pm$ 0.54 & 79.62 $\pm$ 0.55 \\
    RelationNet    & 70.27 $\pm$ 0.63 & 75.69 $\pm$ 0.61 & 75.40 $\pm$ 0.62 \\
    RegressionNet (ours)    & \textbf{72.69} $\pm$ 0.61 & \textbf{80.08} $\pm$ 0.57 & \textbf{80.42} $\pm$ 0.53 \\ 
    \bottomrule
    
  \end{tabular}
  
  \caption{miniImageNet 5-way 10-shot classification accuracies with 95\% confidence intervals for different methods and backbones. For each backbone,
the best method is highlighted.}
\label{tab:5way10shotminidepth}
\end{table}

\section{Domain Shift: mini-ImageNet $\rightarrow$ CUB and CUB $\rightarrow$ mini-ImageNet}\label{app:cross}
We provide additional results on how some of the considered few-shot classification methods perform when the test set is increasingly different from the train set. Note that for RegressionNet we do not use the subspace orthogonalization regularizer from Section \ref{sec:orthogonalization}. In this section we perform experiments by permuting the training and testing domain from one dataset to another (denoted as \textit{train} $\rightarrow$ \textit{test}). The mini-ImageNet $\rightarrow$ CUB setting, introduced by \citet{chen2019closer}, represents a coarse-grained train set to fine-grained test set domain-shift. Conversely, we propose to use the inverse scenario CUB $\rightarrow$ mini-ImageNet as well.

In mini-ImageNet $\rightarrow$ CUB, all 100 classes from mini-ImageNet make up the training set and the same 50 validation and 50 test classes from CUB are used. Conversely, in CUB $\rightarrow$ mini-ImageNet, all 200 classes from CUB make up the training set and the same 16 validation and 20 test classes from mini-ImageNet are used. To illustrate the domain shift, we can look at the mini-ImageNet $\rightarrow$ CUB setting: the 200 classes of CUB are all types of birds, whereas the 64 classes in the training set of mini-ImageNet only contain 3 types of birds, which are not to be found in CUB. Evaluating the domain-shift scenario enables us to understand better which method is more general.

\begin{table}
  \centering
  \begin{adjustbox}{width=\columnwidth,center}
  \begin{tabular}{lllll}
    \toprule
       training set  & CUB     & mini-ImageNet & mini-ImageNet & CUB\\
       test set  & CUB     & mini-ImageNet & CUB &mini-ImageNet\\
    \midrule
    MatchingNet      & 83.75 $\pm$ 0.60 & 69.14 $\pm$ 0.69 & 52.59 $\pm$ 0.71 & 48.95 $\pm$ 0.67\\
    ProtoNet       & 85.70 $\pm$ 0.52 & 73.77 $\pm$ 0.64 & 59.22 $\pm$ 0.74 & 53.58 $\pm$ 0.73\\
    RelationNet   & 82.67 $\pm$ 0.61 & 69.97 $\pm$ 0.68 & 54.36 $\pm$ 0.71 & 45.27 $\pm$ 0.66\\
    RegressionNet (ours)   & \textbf{87.45} $\pm$ 0.48 & \textbf{74.03} $\pm$ 0.68 &
\textbf{62.71} $\pm$ 0.71 & \textbf{56.66} $\pm$ 0.68\\
    \bottomrule
    
  \end{tabular}
  \end{adjustbox}
  \caption{Identical domain and domain shift accuracies for 5-way 5-shot classification using a ResNet-10 backbone and mini-ImageNet and CUB datasets. The best-performing method is highlighted.}
  \label{tab:5way5shotcross}
\end{table}

Table \ref{tab:5way5shotcross} presents our results in the context of existing state-of-the-art distance-metric learning based methods. All experiments in this setting are conducted with a ResNet-10 backbone on a 5-way 5-shot problem. The intra-domain difference between classes is higher in mini-ImageNet than in CUB \citep{chen2019closer}, which is reflected by a drop of the test accuracies of all methods. It can also be seen that when the domain difference between the training and test stage classes increases (left to right in Table \ref{tab:5way5shotcross}) the performance of all few-shot algorithms lowers.

As expected, the classification accuracy lowers when going from the mini-ImageNet $\rightarrow$ CUB to the CUB $\rightarrow$ mini-ImageNet setting. Interestingly though, it does not decrease substantially. This shows that the different feature embeddings generalize well, at least across these two different datasets. Compared to other distance-metric learning based methods, regression networks show significantly better performance in both the mini-ImageNet $\rightarrow$ CUB setting and CUB $\rightarrow$ mini-ImageNet setting.

\end{document}